# Pelvic floor MRI segmentation based on semi-supervised deep learning

Jianwei Zuo, Fei Feng, Zhuhui Wang, James A. Ashton-Miller, John O.L. Delancey and Jiajia Luo

***Abstract*—** The semantic segmentation of pelvic organs via MRI has important clinical significance. Recently, deep learning-enabled semantic segmentation has facilitated the three-dimensional geometric reconstruction of pelvic floor organs, providing clinicians with accurate and intuitive diagnostic results. However, the task of labeling pelvic floor MRI segmentation, typically performed by clinicians, is labor-intensive and costly, leading to a scarcity of labels. Insufficient segmentation labels limit the precise segmentation and reconstruction of pelvic floor organs. To address these issues, we propose a semi-supervised framework for pelvic organ segmentation. The implementation of this framework comprises two stages. In the first stage, it performs self-supervised pre-training using image restoration tasks. Subsequently, fine-tuning of the self-supervised model is performed, using labeled data to train the segmentation model. In the second stage, the self-supervised segmentation model is used to generate pseudo labels for unlabeled data. Ultimately, both labeled and unlabeled data are utilized in semi-supervised training. Upon evaluation, our method significantly enhances the performance in the semantic segmentation and geometric reconstruction of pelvic organs, Dice coefficient can increase by 2.65% averagely. Especially for organs that are difficult to segment, such as the uterus, the accuracy of semantic segmentation can be improved by up to 3.70%.

## I. INTRODUCTION

THE condition known as Pelvic Organ Prolapse (POP), characterized by the abnormal downward displacement and distortion of one or more organs within the female pelvic floor, can cause serious physical and psychological pain to affected women. Each year, approximately 200,000 women in the United States undergo surgical procedures to correct POP, with the associated costs exceeding $1 billion [1,2]. The primary diagnostic modalities employed to evaluate POP include Magnetic Resonance Imaging (MRI) and ultrasound imaging [3]. Owing to its superior soft tissues contrast, MRI is often used for the precise semantic segmentation of pelvic floor organs. This process of organ segmentation is crucial for the reconstruction of three-dimensional (3D) geometric models, beneficial in simulating POP with finite element analysis, and in formulating surgical strategies [4]. Clinicians traditionally perform manual organ segmentation on MRI, a time-consuming task that increases the workload of clinical workers. In addition, the accuracy of segmentation often depends on the clinicians' expertise, making it challenging to ensure the consistency in manual segmentation [5-9].

With the advent of Deep Learning (DL) methods [10] in the medical field and their significant impact, many computer-aided diagnostic approaches, particularly those based on Convolutional Neural Networks (CNNs) [11], have been successfully applied to the analysis of Magnetic Resonance (MR) images for diagnosis. Examples include semantic segmentation and automated diagnosis for various body parts, such as the brain, lungs, liver, blood vessels, musculoskeletal structures, tumors, the heart, and lesion areas [12-19]. In the field of pelvic floor research, He et al. [20] conducted preliminarily investigations into the application of deep learning algorithms for segmenting the main organs of the pelvic floor in abdominal Computed Tomography (CT) images. Moreover, our previous work has explored various diagnostic aspects using CNNs for pelvic floor MRI, including the classification of prolapse types, organ segmentation, and landmark localization [21-23].

The CNN approach improves image analysis efficiency, particularly in large-scale processing. However, the training data are crucial to CNN performance. First, the robust performance of CNNs depends on large training dataset. Second, for supervised learning, training data must be correctly labeled; otherwise, model performance may suffer. However, there are several challenges for the data labelling. First, collecting large medical imaging datasets, akin to natural images, is challenging. Second, the annotation for medical images is costly and time-consuming. The annotation must be

This work was supported by National Key R&D Program of China (2022YFC2402103), National Natural Science Foundation of China (31870942), Peking University Clinical Medicine Plus X-Young Scholars Project PKU2020LCXQ017 and PKU2021LCXQ028, PKU-Baidu Fund 2020BD039, NIH R01 HD038665 and P50 HD044406. (Jianwei Zuo and Fei Feng contributed equally to this work.) (Corresponding author: Jiajia Luo.)

Jianwei Zuo, Zhuhui Wang and Jiajia Luo are with the Institute of Medical Technology, Peking University Health Science Center, Peking University, Beijing 100191, China, and the Biomedical Engineering Department, Peking University, Beijing 100191, China (e-mails: zuojianwei@stu.pku.edu.cn; wang1772@e.ntu.edu.sg; jiajia.luo@pku.edu.cn)

Fei Feng is with the University of Michigan-Shanghai Jiao Tong University Joint Institute, Shanghai Jiao Tong University, Shanghai 200240, China (e-mail: feifeng@sjtu.edu.cn).

Zhuhui Wang is also with the School of Electrical and Electronic Engineering, Nanyang Technological University, Singapore, 639798.

James A. Ashton-Miller is with the Department of Mechanical Engineering, University of Michigan, Ann Arbor, MI 48109, USA (email: jaam@umich.edu)

John O.L. Delancey is with the Department of Obstetrics and Gynecology, University of Michigan, Ann Arbor, Michigan 48109 USA (e-mail: delancey@umich.edu)

conducted by experienced radiologists, leading to a shortage of qualified individuals. Besides, the tedious and time-consuming nature of annotation further limits the availability of labeled images. In many clinical cases, qualitative assessment is prioritized over quantitative evaluation to expedite diagnosis, resulting in fewer annotations. Therefore, it is essential to devise methods that reduce the annotation burden for CNNs in medical applications.

Several approaches are worth exploring. Transfer learning has been used in various studies to enhance image analysis performance with small datasets [24]. For example, Zhang et al. [25] transferred priors experience from large computer vision datasets to multi-modality medical image datasets. Kumar et al. [26] designed an ensemble model that leverages pre-trained models to extract potential features and classify them. Additionally, employing synthetic data for training is an alternative strategy when real data is scarce. For example, Al Khalil et al. [27] utilized synthetic cardiac magnetic resonance images which aid segmentation network generalization and adaptation, and synthetic images show a strong potential to address limited data and privacy issues. Gao et al. [28] explored the potential of synthesized images in increasing the generalization of X-ray analysis algorithms. However, this technique is limited because it requires simple simulation rules, which are challenging to apply to pelvic MR images due to the complexity and diversity of the pelvic floor organs' structure. Given the prevalence of images without annotation, researchers have investigated using unlabeled data through unsupervised learning methods.

Self-supervised learning, an unsupervised learning approach, involves creating proxy tasks to extract general knowledge from unlabeled data [29-31]. He et al. [32] and Chen et al. [33] independently proposed two self-supervised frameworks based on contrastive learning, named MOCO and SimCLR, respectively. He et al. [34] further proposed masked autoencoders (MAE), a self-supervised learning method that masks random patches of the input image to reconstruct the missing pixels, thereby advancing this field. Inspired by the aforementioned studies, self-supervised learning has yielded numerous advances in medicine, aiding deep learning models in feature extraction without annotated data [35]. Moreover, when some labeled data are available, it can guide the CNN training and facilitate model performance evaluation. This approach, utilizing both labeled data and unlabeled data, constitutes a semi-supervised learning method. Semi-supervised learning method not only reduces the need for labeled data but also enables the model to learn from manual annotations, proving beneficial for medical imaging related applications [36-38]. For example, Huynh et al. [39] used semi-supervised learning method to solve the problem of class imbalance in medical image classification. Zhang et al. [40] combined contrastive and semi-supervised learning, and achieved good segmentation performance on a brain dataset with a limited number of labeled images.

In this work, we aimed to develop a semi-supervised framework for pelvic organ segmentation. Semi-supervised

TABLE I
IMAGING PARAMETERS FOR MR IMAGES

| Parameters | NUM |
|---|---|
| TR (ms) | 2108-4000 |
| TE (ms) | 15-30 |
| ETL | 5-8 |
| FA (deg) | 90 |
| NA | 1-2 |
| In-resolution (mm) | 0.78 |
| Th-resolution (mm) | 5.0 |
| Matrix | 256x256 |

TR: repetition time
TE: echo time
ETL: echo train length
FA: flip angle
NA: number of averages
In-resolution: in-plane resolution
Th-resolution: through-plane resolution

learning was conducted using unlabeled data for self-supervised training and generating pseudo labels for the unlabeled data. Further details are provided in the subsequent section.

## II. METHODS

### A. Dataset

To explore the semi-supervised learning effectiveness, a dataset including 48 MR series was used. Since for each MR series, MR images were scanned from three planes of axial, coronal, and sagittal, there was a total of 4103 images. More information about the scanning parameters is shown in Table 1. Then 16 MR series from total were selected for the annotation and there were 1442 images. Organs of the bladder and the uterus were labeled for the 16 MR series. The MR data were collected from Michigan Pelvic Floor Research Collection with ethics approval using a 3T superconducting magnet (Philips Medical Systems Inc, Bothell, WA, USA).

### B. Semi-supervised learning framework

To take advantage of both labeled data and unlabeled data, a semi-supervised learning framework is designed as in Fig.1. There are two stages in the training framework. In the first stage, CNN 1 will be trained with the image recovery task using the unlabeled MR images. Next, CNN 2 will be trained and it will be initialized with the optimized weights of CNN 1 for segmentation using the labeled MR images. Then the first stage ends and the CNN 2 will be used in the second stage for predicting the pseudo labels of unlabeled images. At the second stage, the training process will take advantage of both the unlabeled data and labeled data using CNN 3. The CNN 3 will be initialized with the optimized weights of CNN 1. During forward propagation, CNN 3 will be used for both labeled data and unlabeled data. During backward propagation, the total loss includes both the supervised loss and the unsupervised loss. And the weights of CNN 3 will be updated based on the total loss. Then these steps are introduced in detail below.

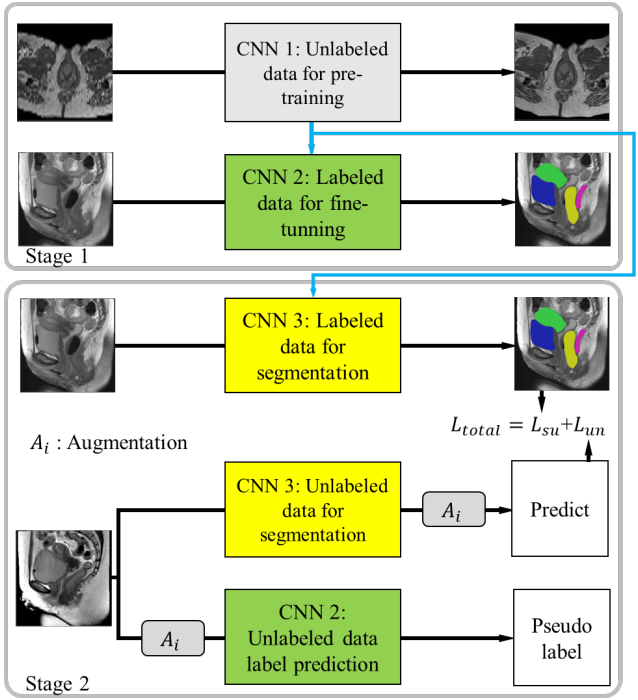

Fig.1. The semi-supervised learning framework for pelvic organ segmentation. Notes: CNN 1-3 share the same model structure but different weights.

### C. Pre-training with unlabeled and labeled data

In this step, the image restoration task is used for self-supervised training and all MR images can be utilized. Degraded images can be created by down sampling high-resolution MR images. Then the CNN 1 can learn the mapping from degraded images to high-resolution images using simulated image pairs. During the simulation, high-resolution MR images are firstly down sampled in one dimension (either row or column dimension), and the image pixels are randomly shuffled along the row or column dimension. The image deterioration can be formulated using the following equation:

$$I^d_{\downarrow r_x S_y}(x, y) = I(x \cdot r_x, S(y)) \quad (1)$$

where $I^d$ is the degraded image and I is the original high-resolution image. $r_x$ represents the sparsity factor for down sampling and $S(\cdot)$ is the shuffling operator. During down sampling, the down sampling ratio will be set to 1:4, 1:6, and 1:8, respectively. Some simulated degraded images are shown in Fig.2.

Then the CNN 1 is trained in a supervised approach. The loss function for CNN 1 is the L1 loss, defined as:

$$Loss = \frac{1}{MN}\sum_{m,n}^{M,N} |g_{mn} - r_{mn}| \quad (2)$$

where M and N are length and width, respectively. $g_{mn}$ and $r_{mn}$ are the pixel values for the high-resolution image and of CNN 1 prediction, on the $m^{th}$ row, $n^{th}$ column, respectively.

Then, the CNN 2 will be trained using labeled data. The loss function for CNN 2 is the cross entropy (CE) loss function [41], which is defined as:

$$CE = \sum_{l=1}^{L}\sum_n -t_{ln}\log(m_{ln}) \quad (3)$$

where L is the total segmentation classes. $t_{ln}$ represents the pixel values of ground truth and $m_{ln}$ represents the segmentation mask for $l^{th}$ class on $n^{th}$ position.

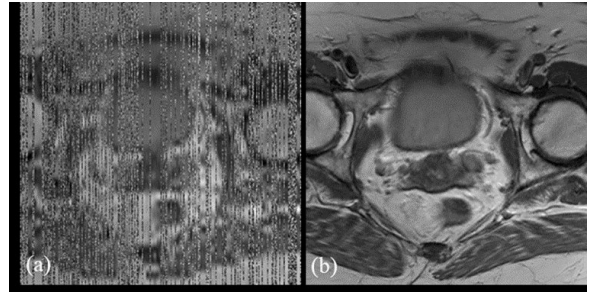

Fig.2. The example of image restoration training pair. (a) the deteriorated image. (b) the raw image.

Besides, the CNN 1 trained in the last step will be used for the model initialization of CNN 2. And the weights of CNN 2 will be fine-tuned based on the labeled data, which ensures that it can have better performance than training from scratch. One thing to note, since the pre-trained weights of CNN 1 will be used for CNN 2 model initialization, they should share the same model structure.

### D. Semi-supervised learning process

In the first stage, CNN 2 is trained for image segmentation using labeled images. It will be used to perform the pseudo-label prediction on unlabeled images. Similarly, the weights of CNN 1 will be used for the initialization of CNN 3, so CNN 3 also has the same model structure as CNN 1. CNN 3 will be used for the semi-supervised image segmentation training. There are two branches, a labeled branch and an unlabeled branch. During training, two branches share the same model structure and weights. The loss function also includes two terms, which is expressed as:

$$L_{total} = L_{su} + L_{un} \quad (4)$$

where $L_{total}$ represents the total loss, $L_{su}$ represents the supervised loss, and $L_{un}$ represents the unsupervised loss. $L_{su}$ is computed by evaluating the difference between prediction and manual annotation. $L_{un}$ is computed by evaluating the difference of prediction and pseudo label predicted by CNN 2. For the labeled data branch, it is also the same as CNN 2's training process, which means it uses MR images with labels and the CE loss function.

However, for the unlabeled branch, it will use the unlabeled MR images as input and use the predictions from CNN 2 as supervision. In this way, it means labels for the unlabeled branch cannot be fully believed. Then, Mean Square Error (MSE) loss [42] is used as the loss function of the unsupervised branch of CNN 3, which is expressed as:

$$MSE = \frac{1}{L*N*M}\sum_{l=1}^{L}\sum_n (t_{ln} - m_{ln})^2 \quad (5)$$

TABLE 2
AVERAGE SELF-SUPERVISED LEARNING PERFORMANCE UNDER DIFFERENT IMAGE RESTORATION TASKS

| Models | SR | PS | Uterus | | Bladder | |
|---|---|---|---|---|---|---|
| | | | Baseline | Self-SL | Baseline | Self-SL |
| U-Net | ✓ | | 0.8155(0.0543) | 0.8275(0.0620) | 0.9002(0.0747) | 0.9102(0.0626) |
| | | ✓ | | 0.7976(0.0736) | | 0.8761(0.0840) |
| | ✓ | ✓ | | **0.8406**(0.0460) | | **0.9146**(0.0620) |
| U-Net++ | ✓ | | 0.7979(0.0403) | 0.8264(0.0520) | 0.9004(0.0658) | 0.9047(0.0782) |
| | | ✓ | | 0.8090(0.0737) | | 0.9051(0.0661) |
| | ✓ | ✓ | | **0.8388**(0.0455) | | 0.9115(0.0673) |

Finally, the segmentation performance is evaluated using Sørensen–Dice coefficient (DSC) [43]:

$$DSC = 2\frac{\sum_n t_{ln} p_{ln}}{\sum_n (t_{ln} + p_{ln})} \quad (6)$$

where $t_{ln}$ represents the pixel values of ground truth and $p_{ln}$ represents the model predicted mask for $l^{th}$ class on $n^{th}$ position.

### III. EXPERIMENTS

To evaluate the effectiveness of the proposed method, labeled MR images were split into three parts for model training, validation, and testing. There are 902 images from ten subjects that were used for training and 540 images from six subjects were used for testing. Among the testing dataset, 180 images for two subjects were used for optimal model validation. Three experiments were conducted to demonstrate the effectiveness of the proposed approach.

First, an ablation study was conducted to investigate the effectiveness of self-supervised learning (CNN 2) and semi-supervised learning (CNN 3) methods. Some experiments were conducted to determine the best training configurations. As for self-supervised learning, the image restoration model achieved both SR and pixel-shuffling (PS) recovery, two tasks were also used individually for comparison. To show the generalization ability of the proposed method, two different classical segmentation CNN models, U-Net [44] and U-Net++ [45] were used. Then, two loss functions of CE and DL were compared during the semi-supervised learning process for U-Net and U-Net++. After determining the optimal training configuration, the author compared the segmentation performances of supervised training from scratch (baseline), self-supervised training, and semi-supervised training methods. Second, performances of models trained with different numbers of training images were quantified to demonstrate the effectiveness of the proposed method for small datasets. Third, the effectiveness of three-view data training was quantified by comparing its performance with the segmentation performances of models trained using only single-view data.

During training, the Adam optimizer was used for training on an NVIDIA TITAN RTX graphics card with 24 GB of computation memory. All models were trained for 800 epochs with a learning rate of 0.0002.

### IV. RESULTS

Different image restoration tasks were tested for the model pre-training and their results were compared in Table 2. The baseline results of U-Net and U-Net++ are used for reference. It shows self-supervised learning training can generally improve the segmentation results compared with baseline model performances. Besides, self-supervised learning training with SR task has better performance than with PS task for both segmentation models. Finally, the proposed image restoration task using both SR and PS obtained the best segmentation performance for both models. For model comparison, U-Net has slightly better performance than U-Net++ in baseline evaluation. However, for self-supervised learning results, the gap between the two models is narrowed.

Then, two different loss functions are compared for the supervised branch of CNN 3 and the results are summarized in Table 3. It shows that semi-supervised learning has better results compared with the baseline results of both models. Besides, it shows that the differences between results obtained from two loss functions are trivial for both models. Then, the author chose to use the CE loss function for the following evaluation.

Next, the segmentation performances of the baseline method, self-supervised learning, and semi-supervised learning are obtained under the different number of images in Table 4 and more intuitive result is shown in Fig.3.

It shows that semi-supervised learning has better segmentation results than baseline and self-supervised learning methods when using the same number of images. Besides, when using data from six subjects for semi-supervised learning, its performance is almost close to the baseline result using data from ten subjects. Similarly, self-supervised learning always has better segmentation performance than the baseline method when using the same number of images. The segmentation performance of all three methods increases as the number of training images increases. However, the segmentation performance gap between semi-supervised learning and baseline methods decrease as the number of training images increase.

TABLE 3
AVERAGE SEMI-SUPERVISED LEARNING PERFORMANCE UNDER DIFFERENT LOSS FUNCTIONS

| Models | Loss | Uterus | | Bladder | |
|---|---|---|---|---|---|
| | | Baseline | Self-SL | Baseline | Self-SL |
| U-Net | DL | | 0.8509(0.0423) | | 0.9122(0.0744) |
| | CE | 0.8155(0.0543) | 0.8445(0.0436) | 0.9002(0.0747) | **0.9235**(0.0534) |
| U-Net++ | DL | | 0.8416(0.0554) | | 0.9158(0.0747) |
| | CE | 0.7979(0.0403) | 0.8488(0.0353) | 0.9004(0.0658) | **0.9169**(0.0602) |

TABLE 4
AVERAGE SEGMENTATION PERFORMANCE WITH THE DIFFERENT NUMBER OF TRAINING IMAGES FOR U-NET

| #of Images /Subjects | Uterus | | | Bladder | | |
|---|---|---|---|---|---|---|
| | Baseline | Self-SL | Semi-SL | Baseline | Self-SL | Semi-SL |
| 182 / 2 | 0.5174(0.2017) | 0.5273(0.1991) | 0.5952(0.1785) | 0.7494(0.1750) | 0.8073(0.1364) | 0.8049(0.1476) |
| 362 / 4 | 0.6197(0.1760) | 0.7106(0.1211) | 0.7247(0.1289) | 0.8086(0.1079) | 0.8535(0.1183) | 0.8675(0.1098) |
| 542 / 6 | 0.7446(0.1003) | 0.7736(0.0746) | 0.8117(0.0668) | 0.8646(0.0772) | 0.8884(0.0838) | 0.9000(0.0589) |
| 722 / 8 | 0.7719(0.0538) | 0.7889(0.0844) | 0.8285(0.0557) | 0.8800(0.0675) | 0.8933(0.0872) | 0.8980(0.0806) |
| 902 / 10 | **0.8155**(0.0543) | **0.8406**(0.0460) | **0.8445**(0.0436) | **0.9002**(0.0747) | **0.9146**(0.0620) | **0.9235**(0.0534) |

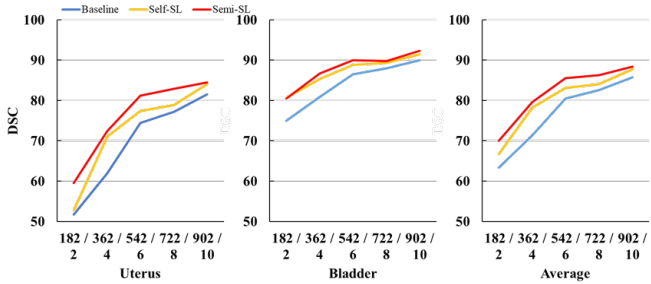

Fig.3. Segmentation performance with the different number of training images under three conditions: baseline (blue), self-supervised learning (yellow), semi-supervised learning (red).

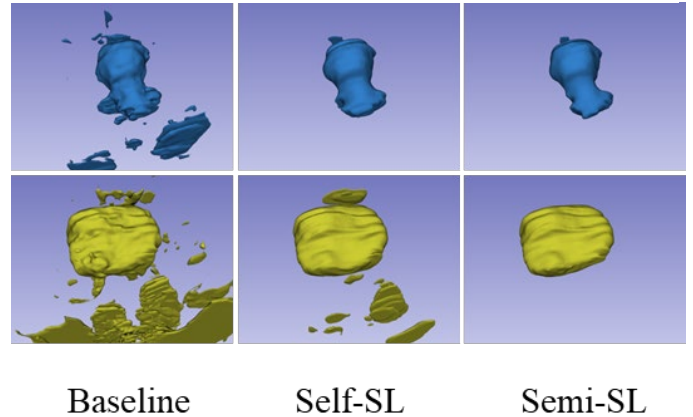

Fig.4. 3D reconstruction performance of uterus(blue) and bladder(yellow) under three conditions: baseline, self-supervised learning and semi-supervised learning.

To further demonstrate the effectiveness of semi-supervised learning method, we performed 3D reconstruction on the segmentation results under the three different conditions in 3D Slicer, as shown in Fig.4. It can be seen that the reconstructed results under the baseline method have significant errors, which are not consistent with the real situation. The reconstruction results under self-supervised learning and semi-supervised learning methods are smoother and closer to real organs. In the three-dimensional reconstruction of the bladder, our proposed semi-supervised learning method also achieved better reconstruction results than the self-supervised learning method.

For a single organ, the bladder always gives better results than the uterus when using the same number of images. Besides, the segmentation performance of the bladder is not bad even when using data from two subjects and the segmentation performance does not increase evidently when using data from six subjects. In contrast, the segmentation performance of the uterus increases evidently with the increase of images.

To investigate whether training segmentation model using three-view data is better than training with single-view data, U-Net was trained using coronal (U-Net$_c$), sagittal (UNet$_s$), and axial (U-Net$_a$) view data. As training using three-view data (U-Net$_{all}$) triples the number of images compared with training using single-view data, the author sampled one in third of images from three-view MR images randomly to train another model (U-Net$_{partial}$). The evaluation results on three-view data are shown in Table 5.

TABLE 5
SEGMENTATION PERFORMANCE COMPARISON AMONG MODELS TRAINED WITH DIFFERENT VIEWS DATA

| Models | Uterus | | | Bladder | | |
|---|---|---|---|---|---|---|
| | Baseline | Self-SL | Semi-SL | Baseline | Self-SL | Semi-SL |
| U-Net$_c$ | 0.5456 (0.2670) | 0.6386 (0.2196) | 0.6240 (0.2332) | 0.7963 (0.1431) | 0.8642 (0.1172) | 0.8812 (0.1045) |
| U-Net$_s$ | 0.5906 (0.2321) | 0.6137 (0.2005) | 0.6478 (0.2125) | 0.8295 (0.1725) | 0.8438 (0.1260) | 0.8671 (0.0874) |
| U-Net$_a$ | 0.5195 (0.2709) | 0.6083 (0.2413) | 0.6168 (0.2752) | 0.6641 (0.2565) | 0.7678 (0.2001) | 0.7831 (0.2320) |
| U-Net$_{partial}$ | 0.7911 (0.0839) | 0.7969 (0.0600) | 0.8170 (0.0691) | 0.8716 (0.1111) | 0.8945 (0.0812) | 0.8982 (0.0912) |
| U-Net$_{all}$ | **0.8155** (0.0543) | **0.8406** (0.0460) | **0.8445** (0.0436) | **0.9002** (0.0747) | **0.9146** (0.0620) | **0.9235** (0.0534) |

It shows that self-supervised learning and semi-supervised learning improve the segmentation results for U-Net$_c$, U-Net$_s$, U-Net$_a$, and U-Net$_{partial}$. Although semi-supervised learning always obtains the best results, the improvements from baseline results to self-supervised learning results are always larger than from self-supervised learning results to semi-supervised learning results for different models. Among these models, U-Net$_{all}$ has the best performance while U-Net$_{partial}$ also has better segmentation performance than U-Net$_c$, U-Net$_s$, and U-Net$_a$.

For a single organ, the bladder always has better results than the uterus for the same model. Besides, the segmentation performance of the bladder is not low even when using single-view data.

Although training with single-view data does not perform well on the data from other views, this does not mean that the model trained with three-view data performs better than the model trained with single-view data for segmentation on this view. Then the segmentation results are compared for each view between U-Net$_{all}$ and U-Net trained with single-view data as shown in Table 6. It shows that for semi-supervised learning, U-Net$_{all}$ has better performances than models trained with only single-view data for all three views.

To demonstrate the effectiveness of the proposed method, the author compared it with the Transfer Learning (TL) method. Segmentation results of transfer learning are even worse than the baseline results, as shown in Table 7.

## V. DISCUSSION

A novel semi-supervised learning method was proposed to improve the performance of CNN-based pelvic organ segmentation. Three novel aspects are featured in this method. First, the intrinsic features of unlabeled MR images are extracted by the CNN model by introducing an image restoration task. Second, a mixed training using both labeled data and unlabeled data is conducted by introducing the hybrid loss and pseudo label. Third, it proves that a CNN model trained with three-view data can produce better performance than models trained with single-view data. The effectiveness of the proposed method was demonstrated through detailed experiments.

### A. Effectiveness of self-supervised tasks

In Table 2, different image restoration tasks were compared for self-supervised learning with two segmentation models. Self-supervised learning has better performances than the baseline method, which proves the effectiveness of self-supervised learning and the image restoration strategy. It indicates that for unlabeled training data, although they could not be used for training the segmentation model directly, it is still feasible to mine the knowledge with self-supervised learning. The mined knowledge is also useful to improve the performance of segmentation. However, there are differences among models obtained with different image restoration tasks.

The image restoration task combining the SR and the PS provides the best performance, which indicates that severer deterioration may help the model learn more general knowledge. Besides, compared to the PS task, the SR task shows better performance, which suggests that the SR task is more related to image segmentation.

Moreover, self-supervised learning is effective for both U-Net and U-Net++, which also suggests the generalization ability of the proposed method. Although there are differences among segmentation performances of the baseline method, the differences are reduced after the self-supervised learning for two CNNs, which suggests that by taking advantage of unlabeled data, the proposed approach can reduce the reliance on model structures.

### B. Effectiveness of loss function

In Table 3, semi-supervised learning was compared with self-supervised learning and the baseline method using two CNNs. And two different loss functions were used for the supervised branch of CNN 3. Semi-supervised learning obtains better results compared with self-supervised learning and the baseline method for two CNNs. It suggests that generating pseudo labels with model predictions is effective to improve the segmentation performance. Although self-supervised learning can learn general knowledge from the image restoration task, some segmentation-specific features may not be captured in this process. It suggests that more segmentation-specific features are learned in the semi-supervised learning process compared to the self-supervised learning process. However, since the predicted segmentation accuracy cannot be known or estimated, the pseudo label cannot be fully trusted. Therefore, the unsupervised branch used the MSE to minimize the loss between the model predictions and the pseudo labels. Since the model trained from self-supervised learning is already an optimized model, the semi-supervised learning model should not be worse than the self-supervised learning model if the model trained with semi-supervised learning produces results similar to the self-supervised learning results on unsupervised data. Besides, since the semi-supervised learning model takes advantage of both the labeled and unlabeled data, it provides the model an opportunity to achieve better segmentation performance than the self-supervised learning model.

TABLE 6
SEMI-SUPERVISED LEARNING SEGMENTATION PERFORMANCE COMPARISON AMONG MODELS TRAINED WITH DIFFERENT VIEWS DATA

| Models | Uterus | | | Bladder | | |
|---|---|---|---|---|---|---|
| | Coronal | Sagittal | Axial | Coronal | Sagittal | Axial |
| U-Net$_c$ | 0.7937 (0.0516) | - | - | 0.9001 (0.0714) | - | - |
| U-Net$_s$ | - | 0.8447 (0.0416) | - | - | 0.9290 (0.0404) | - |
| U-Net$_a$ | - | - | 0.8079 (0.0640) | - | - | 0.8722 (0.0948) |
| U-Net$_{all}$ | **0.8370** (0.0227) | **0.8747** (0.0339) | **0.8218** (0.0205) | **0.9166** (0.0644) | **0.9413** (0.0209) | **0.9124** (0.0591) |

TABLE 7
COMPARISON WITH TRANSFER LEARNING

| Methods | Uterus | Bladder |
|---|---|---|
| Baseline | 0.8155(0.0543) | 0.9002(0.0747) |
| TL | 0.8031(0.0856) | 0.8960(0.0932) |
| Semi-SL | **0.8406**(0.0460) | **0.9146**(0.0620) |

### C. Ablation study for training dataset

In Table 4, the effectiveness of self-supervised learning and semi-supervised learning is illustrated by comparison with the baseline method when using different numbers of images. Semi-supervised learning has the best performance compared to the other two methods when using the same number of training images, which proves the good generalization of the proposed method. Besides, the segmentation performance of semi-supervised learning when trained with data from six subjects is better than that of the baseline method when trained with data from ten subjects, which also suggests that the proposed method can reduce the annotation burden at the cost of unlabeled data. This is because the unlabeled data will be more readily available compared to the manually labeled ones. In addition, semi-supervised learning is more useful to improve the performance when compared with the baseline method, and the improvements from baseline to semi-supervised learning in DSC decrease with the number of training images increases. It suggests that with the increase of labeled data, it becomes more difficult to improve the segmentation performance since the current performance is already pretty good so it requires more training images to increase the performance further. Given this, semi-supervised learning will be more practical because all that need is to collect more MR images. And a better segmentation performance makes the method more practical and useful in clinical practice.

### D. Compared with transfer learning

Furthermore, in Table 7, the proposed approach is also compared with the transfer learning method. The semi-supervised learning method has a better result than the transfer learning result. In previous pelvic organs segmentation works, they found that transfer learning is effective to improve segmentation performance. Such phenomenon can be explained from two aspects. First, heart segmentation is different from pelvic organ segmentation. Second, the number of training images is already large enough and the number of training images of the cardiac MR dataset is not much larger than that of the pelvic MR dataset. In this work, when the number of training images increases to 902 images, the segmentation performance does not improve quickly with the increase of training images. Besides, 902 images are more than three-time bigger than the dataset used in the published papers before (256 images). Therefore, it is difficult to apply transfer learning without a large annotated dataset.

### E. Importance of self-supervised learning

In the proposed method, the importance of self-supervised learning is presented in two aspects. First, it is used for weight initialization in the semi-supervised learning model. Second, it is used to predict the pseudo labels. If the pseudo labels are more accurate, it will be more helpful to improve the final performance. Since self-supervised learning has a better performance than the baseline method, it is more suitable to produce the pseudo labels. As semi-supervised learning can always produce better results than self-supervised learning, which means more useful knowledge has been mined during semi-supervised learning. However, when there are limited annotated images, self-supervised learning does not have a good performance so the pseudo labels are not accurate enough, which also limits the performance of the semi-supervised

48learning. Furthermore, during the semi-supervised learning process, the optimized weights of CNN 1 are used instead of CNN 2. It is because the validation set has been used for the optimization of CNN 2, which may cause data leakage if CNN 2 is used for the initialization of CNN 3. But the data leakage problem does not exist for CNN 1, since it learns different task than segmentation and does not use the segmented data in the optimization process.

### F. Effectiveness of different views

Next, the advantages of training with three view images are investigated. On the one hand, U-Net$_{all}$ can achieve the organ segmentation for three-view data while the single-view models (U-Net$_c$, U-Net$_s$, and U-Net$_a$) can only handle the single-view segmentation problem (Table 5). Therefore, single-view models produce worse results when used for three-view segmentation, even when compared with U-Net$_{partial}$. On the other hand, three-view training can increase the model performance as it can learn the complementary features from different viewpoints. It shows that U-Net$_{all}$ has better performance than three models trained for coronal, sagittal, and axial view segmentation (Table 6). Moreover, U-Net$_{partial}$ (trained with 301 images from ten subjects) has only slightly lower performance than U-Net$_{all}$'s performance (trained with 902 images from ten subjects). It suggests that images from more subjects will be more representative, so choosing images from more subjects for annotation will reduce the annotation burden, and it is more effective than annotating all images from fewer subjects.

### G. Organ comparison

For the segmentation of a single organ, the bladder always obtained a better segmentation performance than the uterus. On the one hand, the reason is that the contour edge of the bladder is clearer and the contrast with surrounding tissues is more pronounced under magnetic resonance imaging. On the other hand, the number of samples in the bladder in the training dataset is bigger than that in the uterus, which helps the model achieve better performance. The bladder can have a good segmentation performance even when training with data from two subjects, while the uterus cannot obtain a good segmentation performance when the number of training images is few. Therefore, when the number of training images increases, the uterus has a greater absolute improvement than the bladder. Besides, there are also differences among different views for the same organ segmentation. In Table 5, it shows bladder has good performances even when training with the single-view data while testing on three-view images. It means that the bladder has high similarities among the three views. For the uterus, the segmentation performances decrease dramatically when using single-view data, which suggests the uterus has different profiles and lower similarities among the three views.

### H. Limitation and future work

Finally, it is worth discussing some prerequisites and unresolved issues of the method. Since it is a semi-supervised learning method, it requires both the labeled and the unlabeled dataset. Besides, the unlabeled dataset should be also larger than the labeled dataset as it needs to mine the useful information from the "big data". Moreover, since pseudo labels will be predicted for the semi-supervised learning training, it requires the unlabeled dataset and labeled dataset having the same data distribution. Some future directions are also needed to explore. The proposed method is successfully applied to the two-organ segmentation problem for MR images scanned at rest. It will make the model more practical if more organs and dynamic MR images can be included. Furthermore, the number of training images is also crucial to the final segmentation performance. More training images from more subjects should be used to improve the segmentation performance. As mentioned above, it is more efficient to annotate partial images from more subjects. Moreover, it is also important to explore the use of MR images from multiple sources. Images from a single hospital or medical center will always be limited as they can only reflect the limited features of the entire population. However, there are some difficulties when dealing with MR images from multiple sources. When the method is applied to data from different distributions, it requires there are labeled images from all distributions. If the labeled data and unlabeled data have a different distribution, the pseudo labels for unlabeled data will be unreliable and may even bring false information, impairing the segmentation performance. Two possible solutions can be explored. The first choice is intuitive that another model can be trained to transfer the unlabeled data to the labeled data domain with methods such as GAN. But it is difficult to control the quality of unpaired data image domain translation. The second choice is that the segmentation model for pseudo label prediction should adapt to different distributions. Previous work has investigated the domain adaption problem of medical image segmentation, which may also be more flexible to deal with images from different sources.

## VI. Conclusion

In this paper, a semi-supervised learning algorithm was devised to improve the performance of pelvic organ segmentation by using both labeled and unlabeled data. The unlabeled data was used in both self-supervised learning and semi-supervised learning. To fully take advantage of the unlabeled data, an image restoration task was designed to pre-train the model before training it with labeled data. Then a hybrid loss was used to combine both the labeled data and unlabeled data for mixed training. From the experimental validation, for different CNNs, the proposed method obtains better results than transfer learning and supervised training with labeled data only, which is effective for smaller datasets. It provides a solution to improve the segmentation performance of pelvic organs without increasing the annotation burden, which may also be beneficial for other MR and CT image segmentation problems.